\documentclass[pdftex]{article}

\usepackage[final, nonatbib]{neurips_2023}
\usepackage[square,sort,comma,numbers]{natbib}
\usepackage{lipsum}
\usepackage{algorithm}
\usepackage{multirow}
\usepackage{algpseudocode}
\usepackage{amsmath}
\usepackage{amssymb}
\usepackage{natbib}
\usepackage{soul}
\usepackage{algorithm}
\usepackage{adjustbox}

\usepackage{wrapfig}
\usepackage[table]{xcolor}
\newcommand{\jygl}{\cellcolor{gray!15}}

\usepackage{hyperref}
\usepackage{fontawesome5}

\usepackage{amsmath,amssymb,amsthm}
\usepackage{graphicx,booktabs,multirow,array}
\usepackage{xcolor}

\usepackage{wrapfig,adjustbox,enumitem,pifont,colortbl,placeins,float}
\definecolor{ourspeach}{HTML}{F8E6D8}

\usepackage[utf8]{inputenc} 
\usepackage[T1]{fontenc}    
\usepackage{hyperref}       
\usepackage{url}            
\usepackage{booktabs}       
\usepackage{amsfonts}       
\usepackage{nicefrac}       
\usepackage{microtype}      
\usepackage{xcolor}         
\usepackage[pdftex]{graphicx} 
\usepackage{microtype}
\usepackage{subfigure}
\usepackage{booktabs} 
\usepackage{color,colortbl}
\usepackage[linewidth=1pt]{mdframed}
\usepackage{framed}
\usepackage{adjustbox}

\usepackage{amsmath}
\usepackage{amssymb}
\usepackage{mathtools}
\usepackage{amsthm}

\usepackage[capitalize,noabbrev]{cleveref}

\theoremstyle{plain}

\theoremstyle{definition}

\theoremstyle{remark}

\usepackage[textsize=tiny]{todonotes}

\title{\Large{Output-Aware Rotation for INT2 KV-Cache Quantization}}

\author{
  Vincent-Daniel Yun\textsuperscript{1,\dag}, 
  Woosang Lim\textsuperscript{2,\dag}, 
  Minsoo Cheong\textsuperscript{2}, 
  Sunwoo Lee\textsuperscript{3} \\ 
  \textbf{
  Murali Annavaram\textsuperscript{1},
  Sai Praneeth Karimireddy\textsuperscript{1},
  Sungjoo Yoo\textsuperscript{2}\thanks{Corresponding Author: sungjoo.yoo@gmail.com.}
    } \\ \\
  \textsuperscript{1}University of Southern California\\ 
  \{yunjuyou, annavara, karimire\}@usc.edu \\
  \textsuperscript{2}Seoul National University \\
  \{ftyg656512, icycle0409\}@snu.ac.kr \\
  \textsuperscript{3}Inha University \\
  \{sunwool\}@inha.ac.kr  \\ \\
  \dag Equal Contribution 
  \vspace{-0.5cm}
}

\begin{document}

\maketitle

\let\thefootnote\relax\footnotetext{}

\begin{abstract}
The key-value (KV) cache has become a major memory and bandwidth bottleneck in long-context large language model inference, making ultra-low-bit quantization increasingly important. However, existing rotation-based INT2 methods optimize cache statistics or proxy errors before the complete attention readout, even though the model is ultimately affected by the error propagated through attention and the output projection $W_O$. To address this mismatch, we propose \textit{OptR}, an output-aware rotation method that minimizes post-$W_O$ attention-output error. OptR decomposes the post-$W_O$ attention-output error into key- and value-induced terms and learns per-head orthogonal corrections through the full INT2 quantization and attention path. OptR further applies an attention-equivalent key reparameterization to reduce large channel-wise offsets without changing the softmax distribution. Across three models and five reasoning and coding benchmarks, OptR consistently improves both QuaRot and OSCAR and strengthens long-context retrieval, while preserving the paged KV-cache format with negligible inference overhead.
\end{abstract}

\begin{center}
\href{https://github.com/daniel-eai/Output-Aware-INT2-KV-Cache-Quantization}
{\faGithub\ GitHub}
\end{center}
\vspace{-0.5cm}

\section{Introduction}

As large language models (LLMs) grow in model size and context length,
the key-value (KV) cache becomes a major bottleneck in long-context
inference~\citep{intro1,gqa}. During autoregressive decoding, each layer
stores the keys and values of all previous tokens and reads them at
every generation step. As a result, KV-cache storage and memory traffic
increase with context length, batch size, and model depth. KV-cache
quantization reduces these costs by storing the cache at lower
precision. We focus on INT2 because it requires only $1/8$ of the
BF16 KV-cache storage and $1/2$ of that of INT4, enabling longer
contexts or larger batches under the same memory budget. Since all values in a quantization group share one scale, a few large values can expand the range represented by only four INT2 levels. This increases rounding error for most values, while aggressive clipping introduces large errors in the outliers themselves~\citep{h2o,kvquant,OSCAR}.

Rotation-based methods reduce this error by spreading a few extreme
channel values across dimensions, and make the cache easier to quantize.
Given an orthogonal matrix $R$, a cache vector $z$ is transformed to
$zR$ before quantization and mapped back with $R^\top$ after
dequantization~\citep{QuIP,Quarot}. Rotation preserves the cache shape
and regular memory layout, maintaining compatibility with paged
KV-cache systems and fused decoding kernels
~\citep{pagedattention,sglang,OSCAR}. The main challenge is selecting
$R$. Existing Rotation-based INT2 KV cache pipelines rely on fixed transforms, or
proxy objectives defined before the complete attention
readout~\citep{Quarot,RotateKV,OSCAR}.

\begin{wrapfigure}{r}{0.5\textwidth}
    \centering
    \includegraphics[width=\linewidth]{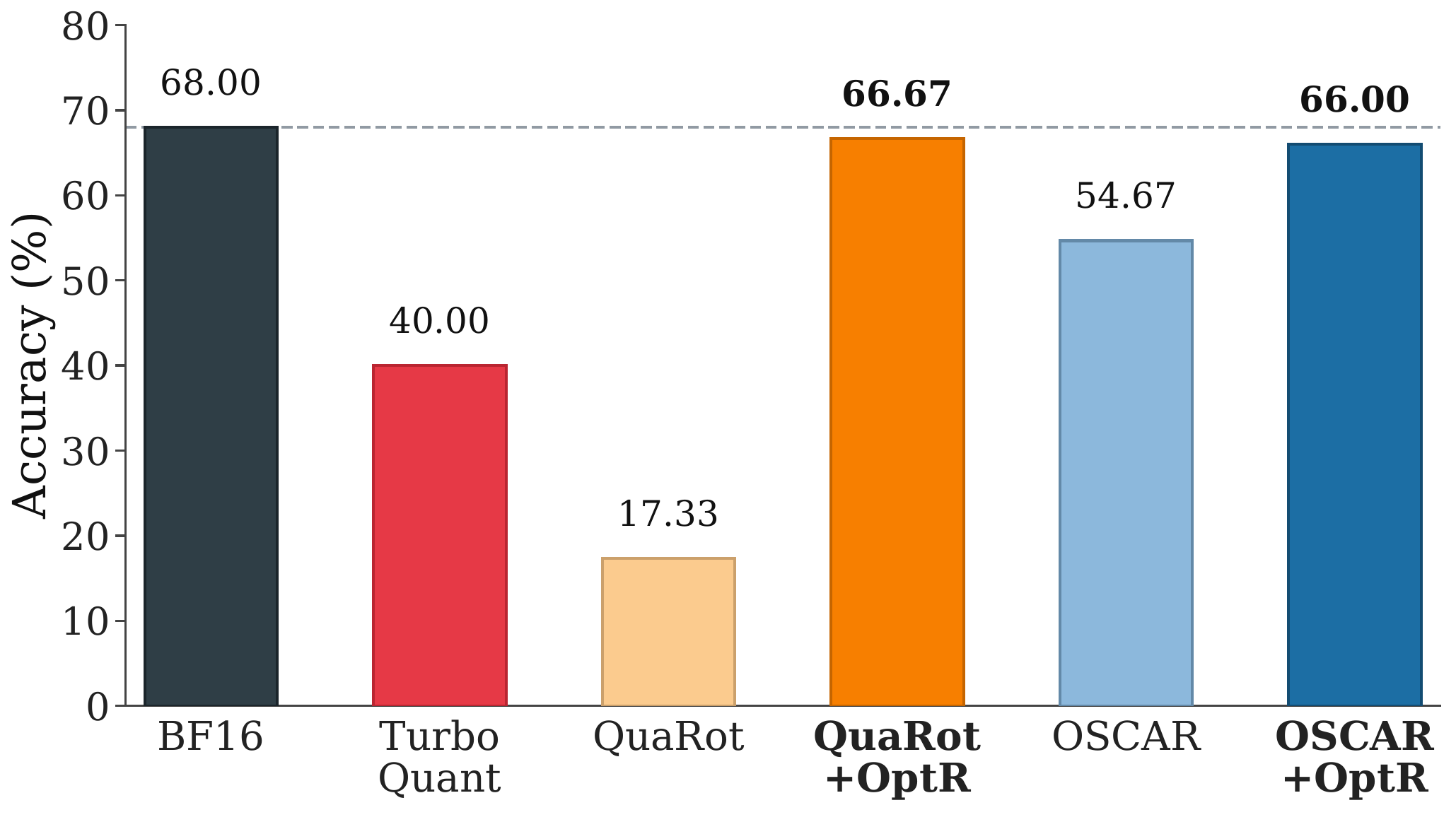}
    \caption{AIME25 accuracy of Qwen3-8B under BF16 and INT2
    KV-cache quantization. $+$ OptR denotes applying key
    reparameterization and output-aware rotation correction to the
    corresponding base rotation. The dashed line marks the BF16
    accuracy.}
    \label{fig:intro_results}
    \vspace{-0.3cm}
\end{wrapfigure}


However, these proxy objectives do not directly measure the error passed to
later layers. KV quantization changes the attention readout, and the
output projection $W_O$ maps this change into the model hidden space.
The resulting output error enters the residual stream and propagates
through subsequent layers, potentially affecting the final prediction.
Consequently, the rotation that best reconstructs the cached keys and
values may not be the one that best preserves the post-$W_O$ output.
This objective mismatch motivates optimizing rotations in output space.

To address this mismatch, we propose \textit{OptR}, an output-aware
rotation method for INT2 KV-cache quantization. OptR first centers the keys before rotation and quantization.
This shifts all logits for a query by the same constant and
therefore leaves the softmax distribution unchanged. Since
INT2 has only four quantization levels, outliers can lead to
large quantization errors. This reparameterization reduces their effect by
narrowing the quantization range. OptR then learns
per-head orthogonal corrections to any base rotation by minimizing the
post-$W_O$ attention-output error through the INT2 attention path. It optimizes
the key rotation before the value rotation because the quantized keys
determine the attention weights used for value aggregation. Only the
rotation parameters are optimized on calibration data. Model weights
remain frozen, and the learned rotations are fixed during inference.

We integrate OptR into an SGLang-based INT2 KV cache pipeline while
retaining paged and prefix-cache support with negligible runtime
overhead~\citep{OSCAR}. Figure~\ref{fig:intro_results} shows that OptR
improves AIME25 accuracy on Qwen3-8B from 17.33\% to 66.67\% with
QuaRot and from 54.67\% to 66.00\% with OSCAR, compared with 68.00\%
for BF16. The gains with both base rotations show that OptR does not
depend on a specific rotation initialization.

Our contributions are summarized as follows.

\begin{itemize}
    \item We formulate INT2 KV-cache quantization as an output-space
    optimization problem and decompose the post-$W_O$ attention-output error
    into key- and value-induced terms.

    \item We propose \textit{OptR}, which first applies
    attention-equivalent key reparameterization to reduce large channel-wise
    offsets and then learns per-head orthogonal corrections through the
    complete INT2 quantization and attention path.


    \item We show that OptR consistently improves existing rotation-based INT2 KV-cache pipelines while retaining their cache layout with negligible serving overhead.
    
\end{itemize}

\section{Related Works}
\paragraph{KV-cache quantization.}
The KV cache grows with context length and is repeatedly read during
decoding, making it a major memory and bandwidth bottleneck. Prior work
reduces this cost through fine-grained quantization, mixed precision,
and vector quantization~\citep{KIVI,kvquant,Kitty,nsnquant}. These designs often introduce
residual buffers, channel-wise metadata, promoted high-precision
channels, or specialized cache layouts, which complicate their
integration with paged KV-cache systems and fused decoding kernels. In
contrast, rotation-based quantization transforms cached vectors into a
quantization-friendly basis without changing their tensor shape or
regular cache layout, making it easier to deploy in existing inference
systems~\citep{Quarot,OSCAR}.

\paragraph{Rotation-based KV-cache quantization.}
QuaRot uses Hadamard rotations for weights, activations, and KV
caches~\citep{Quarot}, while RotateKV adapts rotations to head-specific
key outliers and protects attention sinks~\citep{RotateKV}. OSCAR derives key and value
rotations from offline attention-aware covariance
statistics~\citep{OSCAR}. Despite these differences, existing methods
select rotations using fixed transforms, cache statistics, or proxy
objectives defined before the complete attention readout. Instead, OptR
optimizes rotations against the post-$W_O$ attention-output error produced by the complete INT2 attention path.

\section{Problem Formulation}
\label{sec:problem}

\paragraph{Preliminaries.}
We consider a decoder-only Transformer layer $\ell$ with grouped-query attention (GQA)~\citep{attention, gqa}. Let
$h \in \{1,\ldots,H_{\mathrm{kv}}\}$ denote a KV head, and let
$G_h \subseteq \{1,\ldots,H_q\}$ denote the set of query heads that share this KV head. For
each query head $j \in G_h$, we write
$
q_{t,j}^{\ell},
k_{s,h}^{\ell},
v_{s,h}^{\ell}
\in
\mathbb{R}^{1\times d},
$ where $s \leq t$, $t$ is the current decoding position, and $s$ indexes a cached source token.

Equivalently, the cached keys and values for KV head $h$ up to position $t$ are
$
K_{1:t,h}^{\ell}
=
[k_{1,h}^{\ell};\ldots;k_{t,h}^{\ell}]
\in
\mathbb{R}^{t\times d},
$ and $
V_{1:t,h}^{\ell}
=
[v_{1,h}^{\ell};\ldots;v_{t,h}^{\ell}]
\in
\mathbb{R}^{t\times d}.$
The BF16 attention logits and probabilities are:
\begin{equation}
    a_{t,s}^{\ell,j,h}
    =
    \frac{\langle q_{t,j}^{\ell}, k_{s,h}^{\ell} \rangle}{\sqrt{d}},
    \quad
    p_{t}^{\ell,j,h}
    =
    \operatorname{softmax}_{s \leq t}
    \left(
        a_{t,s}^{\ell,j,h}
    \right)
    \label{eq:fp_attention}
\end{equation}
The corresponding attention output for query head $j$ is given by:
\begin{equation}
    o_{t,j}^{\ell}
    =
    \sum_{s \leq t}
    p_{t,s}^{\ell,j,h}
    v_{s,h}^{\ell}
    =
    (p_{t}^{\ell,j,h})^{\top}
    V_{1:t,h}^{\ell}
    \in
    \mathbb{R}^{1\times d}
    \label{eq:head_output}
\end{equation}
Let $W_{O,j}^{\ell}\in\mathbb{R}^{d_{\mathrm{model}}\times d}$ be the output projection for query head $j$. The attention-output contribution of this head is:
\begin{equation}
    y_{t,j}^{\ell}
    =
    \left(
        \left(p_t^{\ell,j,h}\right)^{\top}
        V_{1:t,h}^{\ell}
    \right)
    (W_{O,j}^{\ell})^{\top}
    \in\mathbb{R}^{1\times d_{\mathrm{model}}}
    \label{eq:residual_readout}
\end{equation}
Thus, the KV cache affects the model through the attention-weighted readout
$(p_t^{\ell,j,h})^{\top}V_{1:t,h}^{\ell}$ and its projection by $W_{O,j}^{\ell}$. 

\begin{figure*}[t]
    \centering
    \includegraphics[width=\textwidth]{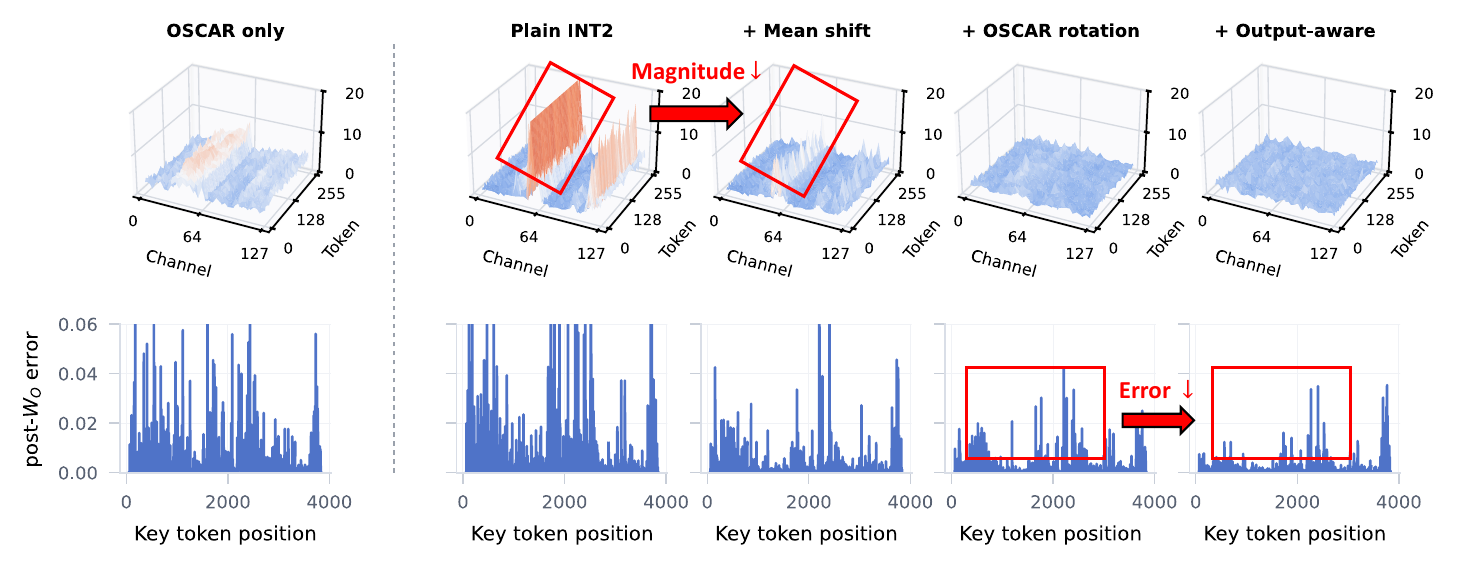}
    \caption{Key magnitude (top) and key-induced attention-output error by cached token (bottom) across five INT2 KV-cache settings on Qwen3-8B (AIME25): OSCAR only, plain INT2, reparameterization (mean shift), reparameterization with OSCAR rotation, and the full OptR pipeline. Lower is better; example details are provided in the Appendix.}
    \label{fig:key_error_motivation}
\end{figure*}

\subsection{Rotated INT2 KV-Cache Quantization}
\label{sec:rotated_int2_kv}

We consider long-context decoding where the long-history KV cache is stored in INT2 and a
small BF16 window is preserved. Let $Q_2(\cdot;c,G)$ denote the INT2
quantize-dequantize map with clipping ratio $c$ and group size $G$. For an orthogonal rotation
$R\in O(d)$, define
\begin{equation}
    D_{R,c}(z)
    =
    Q_2(zR;c,G)R^{\top}
    \label{eq:rotated_int2_map}
\end{equation}
If $Q_2$ is replaced by the identity map, then $D_{R,c}(z)=z$. Thus, the rotation only changes
the coordinate system in which INT2 quantization error is introduced. We denote by
$\widetilde{k}_{s,h}^{\ell}$ and $\widetilde{v}_{s,h}^{\ell}$ the effective keys and values
used by attention after rotated INT2 quantization and BF16 window restoration.

\subsection{Output-Space Error Induced by KV Quantization}
\label{sec:output_distortion_beyond_reconstruction}

We now trace the effective INT2 cache through attention and $W_O$ and
decompose the resulting post-$W_O$ attention-output error into key- and value-induced
terms. With INT2 keys, the attention logits and probabilities become:
\begin{align}
    \widetilde{a}_{t,s}^{\ell,j,h}
    &=
    \frac{
        \langle q_{t,j}^{\ell},\widetilde{k}_{s,h}^{\ell}\rangle
    }{\sqrt{d}},
    \quad
    \widetilde{p}_{t}^{\ell,j,h}
    =
    \operatorname{softmax}_{s\leq t}
    \left(
        \widetilde{a}_{t,s}^{\ell,j,h}
    \right)
    \label{eq:int2_attention}
\end{align}

Let
$\Delta k_{s,h}^{\ell}=\widetilde{k}_{s,h}^{\ell}-k_{s,h}^{\ell}$ and
$\Delta v_{s,h}^{\ell}=\widetilde{v}_{s,h}^{\ell}-v_{s,h}^{\ell}$ denote the key
and value quantization errors. To characterize how key quantization affects
attention, we first express its induced perturbation on the attention logits:
\begin{equation}
    \Delta a_{t,s}^{\ell,j,h}
    =
    \widetilde{a}_{t,s}^{\ell,j,h}
    -
    a_{t,s}^{\ell,j,h}
    =
    \frac{
        \langle q_{t,j}^{\ell}, \Delta k_{s,h}^{\ell} \rangle
    }{\sqrt{d}}
    \label{eq:key_logit_error}
\end{equation}
The resulting attention error is $\Delta p_{t}^{\ell,j,h}
=
\widetilde{p}_{t}^{\ell,j,h}
-
p_{t}^{\ell,j,h}$. Thus, the effect of a key error depends on the query and the softmax attention
map, not only on $\|\Delta k\|_2^2$.
Under INT2 keys and values, the attention-output contribution becomes:
\begin{equation}
    \widetilde{y}_{t,j}^{\ell}
    =
    \Bigl(
        \sum_{s\leq t}
        \widetilde{p}_{t,s}^{\ell,j,h}
        \widetilde{v}_{s,h}^{\ell}
    \Bigr)
    (W_{O,j}^{\ell})^{\top}
    \label{eq:int2_residual_readout}
\end{equation}
Subtracting the BF16 readout gives the exact decomposition. 

Let
$\Delta y_{t,j}^{\ell}:=\widetilde{y}_{t,j}^{\ell}-y_{t,j}^{\ell}$. 
The resulting output error admits an exact decomposition into key-induced and value-induced contributions:
\begin{align}
    \Delta y_{t,j}^{\ell}
    =
    \underbrace{
    \Bigl(
    \sum_{s\leq t}
    \Delta p_{t,s}^{\ell,j,h}v_{s,h}^{\ell}
    \Bigr)
    \Bigl(W_{O,j}^{\ell}\Bigr)^{\top}
    }_{\text{\footnotesize key-induced output error}}
    +
    \underbrace{
    \Bigl(
    \sum_{s\leq t}
    \widetilde{p}_{t,s}^{\ell,j,h}
    \Delta v_{s,h}^{\ell}
    \Bigr)
    \Bigl(W_{O,j}^{\ell}\Bigr)^{\top}
    }_{\text{\footnotesize value-induced output error}}
\end{align}

We denote the two terms above by
$\delta y_{K,t,j}^{\ell,h}$ and $\delta y_{V,t,j}^{\ell,h}$, respectively.
Here $\delta y_{K,t,j}^{\ell,h}$ is the output error induced by key
quantization through the attention distribution, while
$\delta y_{V,t,j}^{\ell,h}$ is the value error after attention-weighted
aggregation and output projection.

This decomposition shows why raw cache reconstruction is only a proxy. A
reconstruction-based objective measures:
\begin{equation}
    E_{\mathrm{rec}}
    =
    \|K-\widetilde{K}\|_{F}^{2}
    +
    \|V-\widetilde{V}\|_{F}^{2}
    \label{eq:raw_reconstruction_objective}
\end{equation}
whereas the model observes the attention-output error:
\begin{equation}
    E_{\mathrm{out}}
    =
    \left\|
        \widetilde{y}_{t,j}^{\ell}
        -
        y_{t,j}^{\ell}
    \right\|_2^2
    =
    \left\|
        \delta y_{K,t,j}^{\ell,h}
        +
        \delta y_{V,t,j}^{\ell,h}
    \right\|_2^2
    \label{eq:output_error_objective}
\end{equation}
Eq.~\eqref{eq:raw_reconstruction_objective} and Eq.~\eqref{eq:output_error_objective} can favor
different rotations because attention and $W_O$ reduce the effect of
some cache errors while allowing others to affect the attention-output. For this reason, OptR uses $E_{\mathrm{out}}$ as its rotation optimization target.

\begin{figure*}[t]
    \centering
    \includegraphics[width=1\textwidth]{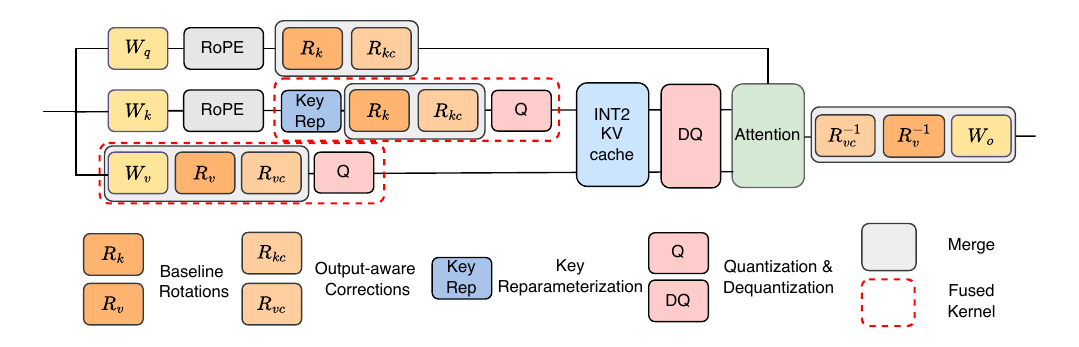}
    \caption{Overview of OptR. OptR augments an existing rotation-based KV-cache quantization pipeline with channel-wise key reparameterization and output-aware correction rotations before INT2 quantization. The resulting quantized KV cache is consumed by attention and projected through $W_O$, where OptR targets reduced post-$W_O$ attention-output error.}
    \label{fig:OptR_overview}
\end{figure*}


\section{Method: Output-Aware Rotation}
\label{sec:OptR}

Figure~\ref{fig:head_heatmap} shows that INT2-induced output error
differs across KV heads. OptR therefore learns a separate orthogonal
correction to the base key and value rotations for each head. During
one-time offline calibration, model weights remain frozen and only the
corrections are optimized to reduce post-$W_O$ attention-output error
through the INT2 attention path. We optimize the key rotation first
because the INT2 keys determine the attention distribution, and then
optimize the value rotation under this distribution. The resulting
rotations are fixed during inference. Figure~\ref{fig:OptR_overview}
summarizes the full pipeline. We omit $(\ell,h)$ when the layer and KV
head are clear.


\subsection{Rotated INT2 Cache Reparameterization}
\label{sec:rotated_int2_cache_reparam}

As shown in Figure~\ref{fig:key_error_motivation}, large channel-wise key offsets increase the dynamic range of group-wise INT2 quantization. Therefore, we reparameterize keys using the per-channel calibration mean $\mu\in\mathbb{R}^{d}$ before rotation and quantization. For key and value rotations $R_K$ and $R_V$, OptR defines
\begin{align}
    \bar{k}_{s}(R_K)
    &=
    D_{R_K,c_K}(k_s-\mu), \quad 
    \bar{v}_{s}(R_V) =D_{R_V,c_V}(v_s) \notag
\end{align}

This key reparameterization is attention-equivalent: subtracting the same $\mu$ from every key adds a query-dependent constant to all logits and leaves the softmax distribution unchanged. We apply it only to keys, since shifting values would alter the attention output. Sink and recent
tokens remain in BF16, and the same centering is applied to their keys. The effective cache is denoted by $\widetilde{k}_{s}(R_K)$ and $\widetilde{v}_{s}(R_V)$.

\subsection{Orthogonal Rotation Refinement}
\label{sec:init_and_correction}

OptR does not rely on a specific rotation initialization.
Let $R_K^0, R_V^0 \in O(d)$ denote arbitrary orthogonal
initializations for the key and value rotations, respectively.
Given these initial rotations, OptR applies the same
output-aware rotation procedure regardless of how they are
constructed.

For each KV head, OptR learns unconstrained matrices
$A_K,A_V\in\mathbb{R}^{d\times d}$ and forms the skew-symmetric
generators:
\begin{equation}
    S_K=A_K-A_K^{\top},
    \quad
    S_V=A_V-A_V^{\top}
    \label{eq:OptR_skew}
\end{equation}
which define the corrected rotations:
\begin{equation}
    R_K(A_K)=R_K^0\exp(S_K),
    \quad
    R_V(A_V)=R_V^0\exp(S_V)
    \notag
\end{equation}
Since $S_K$ and $S_V$ are skew-symmetric, their matrix
exponentials are orthogonal. Therefore, the corrected rotations
remain orthogonal throughout calibration. The same formulation
applies to different rotation initializations without modifying
the calibration objective.

\subsection{Output-Aware Key Calibration}

The key rotation affects attention logits and the attention
distribution next. To isolate this path, we calibrate the key rotation while keeping
values in BF16. The induced key-only attention distribution is:
\begin{align}
    a_{t,s}^{K}(R_K)
    =
    \frac{
        \langle q_{t,j},\widetilde{k}_{s}(R_K)\rangle
    }{\sqrt{d}}, \quad p_t^{K}(R_K)
    =
    \operatorname{softmax}_{s\leq t}
    \Bigl(
        a_{t,s}^{K}(R_K)
    \Bigl)
    \label{eq:OptR_key_attention}
\end{align}
To isolate the effect of key quantization on the attention readout, we define the key-induced attention-output error as:
\begin{equation}
    e_K(t,j;R_K)
    =
    \Bigl[
        \sum_{s\leq t}
        \bigl(
            p_{t,s}^{K}(R_K)-p_{t,s}
        \bigr)
        v_s
    \Bigr]
    (W_{O,j})^{\top}
    \label{eq:key_output_error}
\end{equation}

We then optimize the key rotation to preserve both the attention distribution and the resulting post-$W_O$ output:
\begin{align}
    L_K(R_K;D) =
    \mathbb{E}_{(t,j)\in D}
    \Big[
        D_{\mathrm{KL}}
        \Bigl(
            p_t
            \,\Vert\,
            p_t^{K}(R_K)
        \Bigr)
        +
        \lambda_K
        \frac{
            \|e_K(t,j;R_K)\|_2^2
        }{
            d_{\mathrm{model}}
        }
    \Big]
    \label{eq:OptR_key_loss}
\end{align}
The KL term preserves the attention distribution, while the second term measures the
post-$W_O$ attention-output error caused by key-induced attention changes.

\subsection{Output-Aware Value Calibration}

After optimizing the key rotation, we calibrate the value rotation
under the selected quantized-key attention path. Let $\widehat{R}_K$ be the selected key rotation
and let $\widehat{p}_t^{K}$ be its induced attention distribution.

The value-induced attention-output error is:
\begin{equation}
    e_V(t,j;R_V)
    =
    \Bigl[
        \sum_{s\leq t}
        \widehat{p}_{t,s}^{K}
        \Bigl(
            \widetilde{v}_s(R_V)-v_s
        \Bigl)
    \Bigl]
    (W_{O,j})^{\top}
    \label{eq:value_output_error}
\end{equation}
We optimize the value rotation with:
\begin{equation}
    L_V(R_V;D)
    =
    \mathbb{E}_{(t,j)\in D}
    \Bigl[
        \frac{
            \|e_V(t,j;R_V)\|_2^2
        }{
            d_{\mathrm{model}}
        }
    \Bigl]
    \label{eq:OptR_value_loss}
\end{equation}
This objective measures value-induced attention-output error after attention-weighted aggregation and output projection,
rather than raw value-cache reconstruction.

\subsection{INT2-Aware Orthogonal Calibration}
OptR calibrates the rotation generators through the full INT2 quantization
path: rotation, clipping, grouping, INT2 rounding, dequantization, and inverse
rotation. For each layer and KV head, OptR learns the orthogonal
correction:
\begin{align}
    A_K^{\star}
    =
    \arg\min_A
    L_K
    \Bigl(
        R_K^0\exp(A-A^{\top});
        D_{\mathrm{train}}
    \Bigl),
    \quad
    A_V^{\star}=
    \arg\min_A
    L_V
    \Bigl(
        R_V^0\exp(A-A^{\top});
        D_{\mathrm{train}}
    \Bigl)
    \label{eq:OptR_opt}
\end{align}
All model weights remain frozen; only the per-head rotation generators are calibrated. Since
INT2 rounding is non-differentiable, gradients are estimated using a straight-through estimator:
\begin{equation}
    \frac{\partial Q_2(Z;c,G)}{\partial Z}
    \approx
    \mathbf{1}\{|Z|\leq\tau_c\}
    \label{eq:OptR_ste}
\end{equation}
where $\tau_c$ is the clipping threshold determined by the clip ratio $c$.

\section{Experimental Results}
\noindent\textbf{Models and Benchmarks.}
We evaluate OptR on Qwen3-4B-Thinking-2507~\citep{qwen3}, Qwen3-8B~\citep{qwen3}, and Phi4-14B-reasoning-plus~\citep{phi4reasoning}. We measure reasoning and coding accuracy on AIME24, AIME25~\citep{aime25}, GPQA-Diamond~\citep{gpqa}, MBPP+~\citep{evalplus}, and LiveCodeBench v6~\citep{livecodebench}. We additionally evaluate long-context retrieval using RULER-NIAH (Needle-in-a-Haystack)~\citep{ruler}, with context lengths up to 64K for the Qwen3 models and 32K for Phi4-14B-reasoning-plus. For all models, we use a temperature of 0.6, top-$p$ of 0.95, and top-$k$ of 20. For the reasoning and coding benchmarks, we use maximum generation lengths of 32K for the Qwen3 models and 16K for Phi4-14B-reasoning-plus.


\newcommand{\spacedhline}{%
    \noalign{\vskip 2pt}%
    \hline
    \noalign{\vskip 2pt}%
}

\begin{table*}[t]
\centering
\caption{Comparison of INT2 KV-cache quantization methods across three model configurations and five benchmarks. Results are reported as $\mu \pm \sigma$ over five seeds. OptR indicates that output-aware rotation is applied to the corresponding baseline. BPE denotes the effective number of bits per KV-cache element. TurboQuant is reported from a single run because repeated 32K evaluations are slow in its vLLM implementation.}
\vspace{0.2cm}
\scriptsize
\setlength{\tabcolsep}{2pt}
\renewcommand{\arraystretch}{1.2}

\resizebox{\textwidth}{!}{
\begin{tabular}{llccccccc}
\hline
\textbf{Model} & \textbf{Method} & \textbf{BPE} & \textbf{AIME24} & \textbf{AIME25} & \textbf{GPQA} & \textbf{MBPP$+$} & \textbf{LCB v6} & \textbf{Mean} \\
\spacedhline

\multirow{6}{*}{\begin{tabular}[c]{@{}l@{}}Qwen3-4B-\\Thinking-2507\end{tabular}}
& BF16 & 16 & $78.00 \pm 6.91$ & $71.33 \pm 3.80$ & $64.55 \pm 2.30$ & $77.31 \pm 0.87$ & $46.71 \pm 2.05$ & $67.58$ \\
& TurboQuant (no MP) & 3.25 & $10.00$ & $16.67$ & $41.92$ & $23.02$ & $1.71$ & $18.66$ \\
& QuaRot-INT2 & 2.32 & $0.00 \pm 0.00$ & $0.00 \pm 0.00$ & $6.06 \pm 1.18$ & $5.61 \pm 2.36$ & $1.37 \pm 0.31$ & $2.61$ \\
& \jygl QuaRot-INT2 + \textbf{OptR} & \jygl 2.32 & \jygl $69.33 \pm 4.35$ & \jygl $60.67 \pm 4.35$ & \jygl $62.12 \pm 2.92$ & \jygl $\mathbf{78.31 \pm 1.23}$ & \jygl $37.71 \pm 0.81$ & \jygl $61.63$ \\
& OSCAR & 2.32 & $68.67 \pm 2.98$ & $63.33 \pm 4.08$ & $62.02 \pm 1.73$ & $75.66 \pm 0.84$ & $43.89 \pm 0.75$ & $62.71$ \\
& \jygl OSCAR + \textbf{OptR} & \jygl 2.32 & \jygl $\mathbf{72.00 \pm 1.83}$ & \jygl $\mathbf{70.67 \pm 2.79}$ & \jygl $\mathbf{62.83 \pm 2.22}$ & \jygl $76.83 \pm 0.91$ & \jygl $\mathbf{44.57 \pm 0.90}$ & \jygl $\mathbf{65.38}$ \\

\spacedhline

\multirow{6}{*}{Qwen3-8B}
& BF16 & 16 & $76.00 \pm 6.41$ & $68.00 \pm 2.98$ & $55.76 \pm 2.16$ & $79.15 \pm 1.48$ & $49.37 \pm 0.65$ & $65.66$ \\
& TurboQuant (no MP) & 3.25 & $53.33$ & $40.00$ & $45.96$ & $58.99$ & $21.14$ & $43.88$ \\
& QuaRot-INT2 & 2.32 & $16.00 \pm 5.48$ & $17.33 \pm 4.35$ & $42.12 \pm 1.54$ & $49.10 \pm 1.00$ & $5.26 \pm 1.24$ & $25.96$ \\
& \jygl QuaRot-INT2 + \textbf{OptR} & \jygl 2.32 & \jygl $76.00 \pm 4.35$ & \jygl $\mathbf{66.67 \pm 3.33}$ & \jygl $57.58 \pm 2.45$ & \jygl $\mathbf{80.63 \pm 1.14}$ & \jygl $39.77 \pm 1.54$ & \jygl $64.13$ \\
& OSCAR & 2.32 & $72.00 \pm 3.80$ & $54.67 \pm 3.80$ & $56.26 \pm 3.25$ & $78.73 \pm 0.44$ & $45.03 \pm 2.26$ & $61.34$ \\
& \jygl OSCAR + \textbf{OptR} & \jygl 2.32 & \jygl $\mathbf{76.67 \pm 2.36}$ & \jygl $66.00 \pm 4.94$ & \jygl $\mathbf{58.99 \pm 2.56}$ & \jygl $79.21 \pm 0.55$ & \jygl $\mathbf{47.54 \pm 0.48}$ & \jygl $\mathbf{65.68}$ \\

\spacedhline

\multirow{6}{*}{\begin{tabular}[c]{@{}l@{}}Phi4-14B-\\reasoning-plus\end{tabular}}
& BF16 & 16 & $70.00 \pm 4.71$ & $60.67 \pm 4.94$ & $45.56 \pm 2.09$ & $77.18 \pm 2.16$ & $39.43 \pm 2.06$ & $58.57$ \\
& TurboQuant (no MP) & 3.25
& $60.00$
& $53.33$
& $46.97$
& $74.60$
& $33.14$
& $53.61$ \\
& QuaRot-INT2 & 2.32 & $61.33 \pm 4.47$ & $46.00 \pm 5.48$ & $46.77 \pm 3.40$ & $75.82 \pm 1.16$ & $34.97 \pm 1.48$ & $52.98$ \\
& \jygl QuaRot-INT2 + \textbf{OptR} & \jygl 2.32 & \jygl $64.00 \pm 3.65$ & \jygl $52.00 \pm 2.98$ & \jygl $\mathbf{47.58 \pm 2.91}$ & \jygl $\mathbf{76.70 \pm 0.67}$ & \jygl $\mathbf{36.00 \pm 0.57}$ & \jygl $55.26$ \\
& OSCAR & 2.32 & $62.67 \pm 2.79$ & $49.33 \pm 3.65$ & $43.84 \pm 2.71$ & $73.02 \pm 1.44$ & $34.29 \pm 1.04$ & $52.63$ \\
& \jygl OSCAR + \textbf{OptR} & \jygl 2.32 & \jygl $\mathbf{65.33 \pm 5.06}$ & \jygl $\mathbf{58.00 \pm 5.58}$ & \jygl $46.16 \pm 1.50$ & \jygl $73.44 \pm 0.55$ & \jygl $35.09 \pm 1.11$ & \jygl $\mathbf{55.60}$ \\

\spacedhline
\end{tabular}
}
\label{tab:int2_kv_quant}
\end{table*}

\begin{table*}[t]
\centering
\caption{RULER-NIAH long-context retrieval accuracy evaluated at context lengths ranging from 4k to 64k tokens. Results are reported as $\mu \pm \sigma$ over three random seeds, with 800 examples evaluated per seed. OptR indicates that output-aware rotation is applied to the corresponding baseline. Since Phi4-14B-reasoning-plus supports a maximum context length of 32k, its evaluation is limited to 32k, and the 64k setting is reported as N/A. Additional 128K results are in the Appendix.}
\vspace{0.2cm}
\scriptsize
\renewcommand{\arraystretch}{1.2}

\begin{adjustbox}{width=1\textwidth}
\begin{tabular}{llccccc}
\hline
\textbf{Model} & \textbf{Method} & \textbf{4k} & \textbf{8k} & \textbf{16k} & \textbf{32k} & \textbf{64k} \\
\spacedhline

\multirow{5}{*}{\begin{tabular}[c]{@{}l@{}}Qwen3-4B-\\Thinking-2507\end{tabular}}
& BF16 & $99.90 \pm 0.08$ & $99.60 \pm 0.07$ & $98.19 \pm 0.08$ & $97.13 \pm 0.11$ & $88.76 \pm 0.47$ \\
& QuaRot-INT2 & $3.08 \pm 0.19$ & $8.26 \pm 0.85$ & $0.00 \pm 0.00$ & $6.14 \pm 1.11$ & $0.72 \pm 0.16$ \\
& \jygl QuaRot-INT2 + \textbf{OptR}
& \jygl $\mathbf{99.78 \pm 0.17}$
& \jygl $\mathbf{99.02 \pm 0.38}$
& \jygl $\mathbf{95.59 \pm 0.52}$
& \jygl ${74.69 \pm 0.99}$
& \jygl ${46.86 \pm 1.77}$ \\
& OSCAR & $99.29 \pm 0.31$ & $97.59 \pm 0.50$ & $93.78 \pm 0.33$ & $77.75 \pm 0.64$ & $55.72 \pm 1.44$ \\
& \jygl OSCAR + \textbf{OptR}
& \jygl ${99.42 \pm 0.27}$
& \jygl ${98.57 \pm 0.17}$
& \jygl ${95.31 \pm 0.10}$
& \jygl $\mathbf{84.98 \pm 0.45}$
& \jygl $\mathbf{69.58 \pm 1.31}$ \\

\spacedhline

\multirow{5}{*}{Qwen3-8B}
& BF16 & $99.83 \pm 0.11$ & $99.93 \pm 0.02$ & $99.45 \pm 0.08$ & $98.70 \pm 0.49$ & $84.22 \pm 1.41$ \\
& QuaRot-INT2 & $84.97 \pm 0.09$ & $41.73 \pm 5.75$ & $18.16 \pm 1.14$ & $13.03 \pm 1.71$ & $0.04 \pm 0.07$ \\
& \jygl QuaRot-INT2 + \textbf{OptR}
& \jygl ${99.40 \pm 0.29}$
& \jygl $\mathbf{98.76 \pm 0.57}$
& \jygl $\mathbf{96.14 \pm 0.35}$
& \jygl ${86.37 \pm 0.72}$
& \jygl $\mathbf{70.02 \pm 1.79}$ \\
& OSCAR & $99.59 \pm 0.15$ & $97.94 \pm 0.28$ & $94.39 \pm 0.29$ & $83.76 \pm 0.58$ & $57.54 \pm 1.82$ \\
& \jygl OSCAR + \textbf{OptR}
& \jygl $\mathbf{99.60 \pm 0.11}$
& \jygl ${98.16 \pm 0.63}$
& \jygl ${95.50 \pm 0.65}$
& \jygl $\mathbf{86.42 \pm 0.55}$
& \jygl ${68.65 \pm 0.80}$ \\

\spacedhline

\multirow{5}{*}{\begin{tabular}[c]{@{}l@{}}Phi4-14B-\\reasoning-plus\end{tabular}}
& BF16 & $98.50 \pm 0.45$ & $97.54 \pm 0.59$ & $97.76 \pm 0.70$ & $92.13 \pm 1.34$ & N/A \\
& QuaRot-INT2 & $96.28 \pm 0.44$ & $92.28 \pm 1.31$ & $84.54 \pm 0.52$ & $69.65 \pm 0.94$ & N/A \\
& \jygl QuaRot-INT2 + \textbf{OptR}
& \jygl $\mathbf{97.62 \pm 0.21}$
& \jygl $\mathbf{94.02 \pm 0.80}$
& \jygl $\mathbf{89.98 \pm 0.52}$
& \jygl $\mathbf{76.21 \pm 0.53}$
& \jygl N/A \\
& OSCAR & $95.62 \pm 1.02$ & $90.50 \pm 0.24$ & $84.59 \pm 0.86$ & $70.09 \pm 0.59$ & N/A \\
& \jygl OSCAR + \textbf{OptR}
& \jygl ${96.30 \pm 0.37}$
& \jygl ${93.42 \pm 0.35}$
& \jygl ${87.64 \pm 0.45}$
& \jygl ${75.36 \pm 0.85}$
& \jygl N/A \\
\spacedhline
\end{tabular}
\end{adjustbox}
\label{tab:ruler_niah}
\vspace{-0.2cm}
\end{table*}

\paragraph{Implementation Details.}OptR is implemented on top of the official OSCAR codebase~\citep{OSCAR}. We use group-wise affine INT2 quantization with a group size of 128, clipping ratios of 0.96 for keys and 0.92 for values, and retain 64 sink tokens and 256 recent tokens in BF16. $\lambda_K$ is fixed to 1.0 across all models and benchmarks.
We report an effective cache cost of 2.32 bits per element (BPE) at a 64K-token context, including quantization metadata and the BF16 windows. For each model, we collect a 30K-token pool of BF16 GPQA QKV traces and use disjoint subsets for calibration and held-out rotation selection. For each layer and KV head, we optimize only the rotation corrections for 80 Adam steps with a learning rate of 0.02~\citep{kingma2014adam}. All model weights remain frozen. All experiments were conducted on four NVIDIA A100 40GB GPUs. Additional implementation details are provided in the Appendix. 

\paragraph{Baselines.}
We compare OptR with BF16, TurboQuant~\citep{TurboQuant} without mixed precision (no MP), QuaRot-INT2~\citep{Quarot}, and OSCAR~\citep{OSCAR}. OptR is applied to both QuaRot and OSCAR to evaluate its effectiveness across fixed Hadamard rotations and the state-of-the-art attention-aware covariance rotation. All rotation-based methods use the same INT2 group size and BF16 sink and recent-token windows.

\subsection{Main Results}
\paragraph{Overall Accuracy.}
Table~\ref{tab:int2_kv_quant} compares OptR with BF16 and existing INT2 KV-cache quantization methods. QuaRot-INT2 often suffers from severe accuracy degradation, particularly on reasoning and coding tasks. Applying OptR substantially improves both QuaRot and OSCAR at the same 2.32 BPE across the evaluated models and benchmarks.

\begin{figure*}[t]
\centering
\includegraphics[width=1\linewidth]{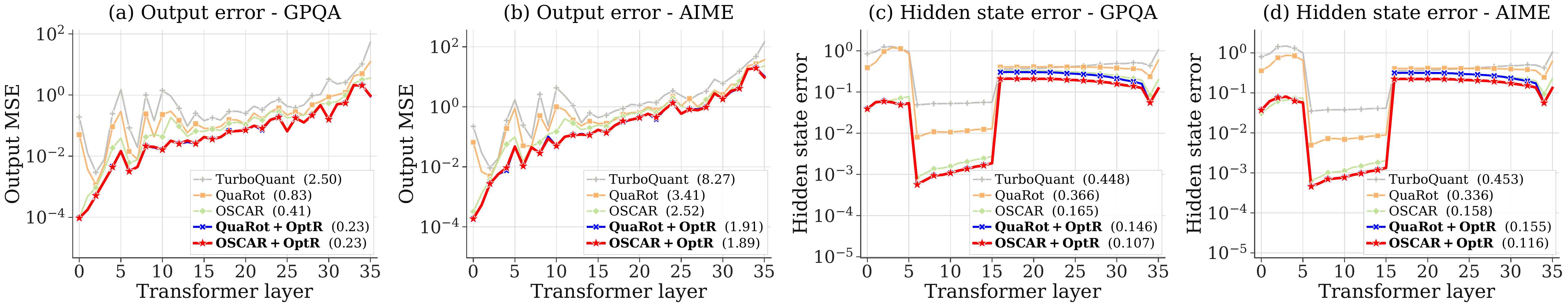}
\caption{Per-layer errors under INT2 KV-cache quantization on Qwen3-4B-Thinking-2507. Panels (a,b) show post-$W_O$ attention-output error on GPQA-Diamond and AIME25 at each layer. Panels (c,d) show the propagated residual-stream error, measured as the normalized squared error between BF16 and INT2-KV block-output hidden states on the same BF16-generated token sequences. Legend values indicate the mean across layers. Error values are shown on a logarithmic scale. Detailed settings are provided in Appendix.}
\label{fig:per-layer-error}
\end{figure*}

\begin{figure*}[t]
\centering
\includegraphics[width=1\textwidth]{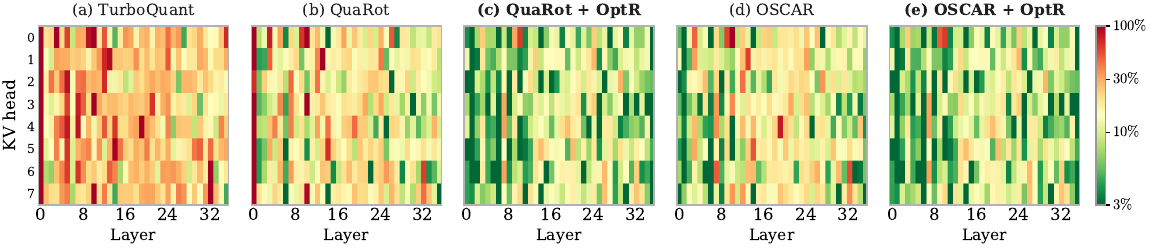}
\caption{Output error by layer and KV head relative to naive INT2
(Qwen3-4B-Thinking-2507, GPQA-Diamond). Each cell reports the post-$W_O$
attention-output RMS error as a percentage of plain per-group INT2 without
rotation. Lower values indicate smaller output error, with naive INT2
corresponding to $100\%$. Panels show (a) TurboQuant, (b) QuaRot,
(c) QuaRot + OptR, (d) OSCAR, and (e) OSCAR + OptR.
Detailed settings are provided in Appendix.}
\label{fig:head_heatmap}
\end{figure*}

\paragraph{Long-Context Robustness.}
Table~\ref{tab:ruler_niah} reports RULER-NIAH ~\citep{ruler} retrieval accuracy across increasing context lengths. QuaRot-INT2 degrades rapidly as the context grows, whereas OptR preserves substantially stronger retrieval performance for both QuaRot and OSCAR. The gains become more pronounced at longer contexts, indicating that OptR more effectively mitigates accumulated KV quantization error over long histories.

\paragraph{Output-Error Analysis.}
Figure~\ref{fig:per-layer-error} shows that OptR reduces both post-$W_O$ attention-output error and propagated residual-stream error relative to QuaRot and OSCAR on GPQA-Diamond and AIME25. Figure~\ref{fig:head_heatmap} further shows that these reductions occur across most layers and KV heads rather than only a small subset.

\begin{table}[h]
\centering
\caption{Component-wise evaluation of OptR on AIME25 across Qwen3-4B-Thinking-2507 and Phi4-14B-reasoning-plus using the state-of-the-art OSCAR rotation.}
\footnotesize
\begin{tabular}{lcc}
\hline
\textbf{Method}
& \begin{tabular}[c]{@{}c@{}}\textbf{Qwen3-4B-}\\\textbf{Thinking-2507}\end{tabular}
& \begin{tabular}[c]{@{}c@{}}\textbf{Phi4-14B-}\\\textbf{reasoning-plus}\end{tabular} \\
\hline
BF16
& $71.33 \pm 3.80$
& $60.67 \pm 4.94$ \\

OSCAR
& $63.33 \pm 4.08$
& $49.33 \pm 3.65$ \\

OSCAR + Key Reparam
& $66.00 \pm 4.71$
& $54.67 \pm 6.91$ \\

\jygl OSCAR + \textbf{OptR}
& \jygl $\mathbf{70.67 \pm 2.79}$
& \jygl $\mathbf{58.00 \pm 5.58}$ \\
\hline
\end{tabular}
\label{tab:optr_ablation}
\end{table}

\begin{table}[h]
\centering
\caption{
Objective ablation for OptR on Qwen3-4B-Thinking-2507 with
the OSCAR base rotation. $\mathcal{E}(X)=\operatorname{mean}(X^2)$ denotes the element-wise mean squared error. The $D_{KL}$ term is fixed across all rows, and the three rows vary the error term using cache reconstruction, pre-$W_O$ attention readout, and post-$W_O$ attention-output error, respectively.
}
\footnotesize
\renewcommand{\arraystretch}{1.35}
\begin{tabular}{ccc}
\hline
\textbf{Key Objective} $\mathcal{L}_K$
& \textbf{Value Objective} $\mathcal{L}_V$
& \textbf{AIME25} \\
\hline

$D_{KL}+\lambda_K\mathcal{E}\!\left(\widetilde K-K\right)$
&
$\mathcal{E}\!\left(\widetilde V-V\right)$
&
$65.33 \pm 6.91$ \\

$D_{KL}+\lambda_K\mathcal{E}\!\left(\Delta p_K^\top V\right)$
&
$\mathcal{E}\!\left((\widehat p_t^K)^\top\Delta V\right)$
&
$62.67 \pm 2.79$ \\

\jygl
$D_{\mathrm{KL}}
+\lambda_K\mathcal{E}\!\left(
    \left(\Delta p_K^\top V\right)
    W_{O,j}^{\top}
\right)$
&
\jygl
$\mathcal{E}\!\left(
    \left((\widehat p_t^K)^\top\Delta V\right)
    W_{O,j}^{\top}
\right)$
&
\jygl $\mathbf{70.67 \pm 2.79}$ \\

\hline
\end{tabular}
\label{tab:objective_ablation}
\end{table}

\begin{figure*}[t]
\centering
\includegraphics[width=1\textwidth]{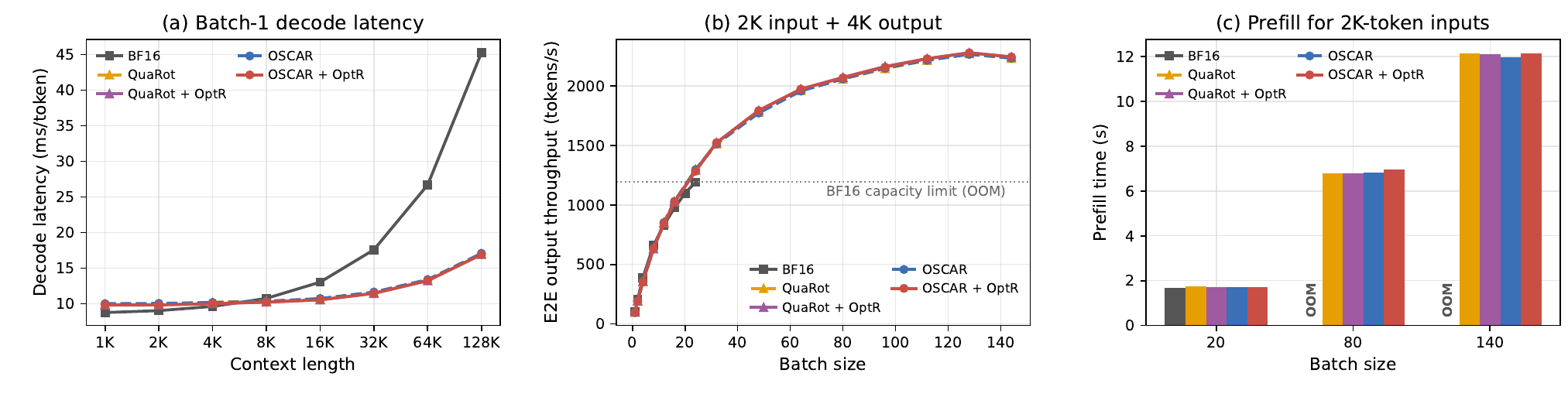}
\caption{Efficiency analysis of OptR integrated into the optimized rotated INT2 KV cache pipeline on an NVIDIA A100 40GB using Qwen3-4B-Thinking-2507. (a) Batch-1 decode latency across context lengths from 1K to 128K. (b) End-to-end output throughput, for a 2K-input and 4K-output workload at different batch sizes. (c) Prefill time for 2K-token inputs at representative batch sizes. Analysis shows that OptR adds negligible inference overhead to the underlying pipeline.}
\label{fig:optr_efficiency}
\end{figure*}

\paragraph{Ablation Study.}
All ablation results are reported as $\mu \pm \sigma$ over five seeds. Table~\ref{tab:objective_ablation} compares cache-reconstruction, pre-$W_O$, and post-$W_O$ calibration objectives on Qwen3-4B-Thinking-2507 using the same OSCAR base rotation. The post-$W_O$ attention-output objective achieves the highest AIME25 accuracy. Table~\ref{tab:optr_ablation} shows that the full OptR consistently achieves the strongest performance across both base rotations and models, with key reparameterization providing an additional source of improvement. Additional sensitivity ablation results for $\lambda_K$ tuning are provided in the Appendix.

\section{Discussion}
\paragraph{Efficiency Analysis.}
We integrate OptR into an optimized INT2 KV-cache pipeline and extend
its Triton cache-write kernel to support per-KV-head key rotations and
key reparameterization. For each head $h$, reparameterization before rotation
can be rewritten as:
\[
    (k-\mu_h)R_{K,h}
    =
    kR_{K,h}-\mu_hR_{K,h}
\]
We therefore precompute the rotated mean $\mu_hR_{K,h}$ once offline
and subtract it immediately after rotating each key. The same kernel
then clips, quantizes, packs, and writes the reparameterized keys and rotated
values to the INT2 cache in a single launch. The value rotation and
its inverse are absorbed into the value projection and $W_O$,
respectively, avoiding separate value-side rotation kernels. To match
the per-KV-head key rotations under GQA, we implement an optimized
Triton kernel that applies the corresponding rotation to each query
head.

Figure~\ref{fig:optr_efficiency} shows that OptR remains within 2\% of
the corresponding base pipeline in decode latency, end-to-end
throughput, and prefill time across the evaluated settings, while
retaining the same maximum batch sizes. For the 2K-token input and 4K-token output workload in Figure~\ref{fig:optr_efficiency}(b), adding OptR at $B=128$ increases GPU memory by only 18 MiB, from 36,205 MiB to 36,223 MiB. Thus, OptR preserves the efficiency and serving capacity of the
underlying INT2 pipeline.




\section{Conclusion}
We introduced OptR, an output-aware rotation method for INT2 KV-cache quantization. OptR combines attention-equivalent key reparameterization with per-head orthogonal corrections optimized through the complete quantized attention path, directly preserving the attention output observed by subsequent layers. Across diverse models, reasoning and coding tasks, and long-context retrieval settings, OptR consistently strengthens existing rotation-based methods and reduces quantization-induced output error. Moreover, OptR retains compatibility with paged KV-cache serving and introduces negligible inference overhead, demonstrating that output-aware optimization is an effective and practical direction for ultra-low-bit KV-cache quantization.


\bibliographystyle{plain}
\bibliography{custom}

\appendix
\setcounter{figure}{0}
\setcounter{table}{0}

\renewcommand{\thefigure}{\Alph{figure}}
\renewcommand{\thetable}{\Alph{table}}

\clearpage
\newpage
\section{Supplementary Materials}
Additional implementation details, experimental settings, and qualitative analyses are provided in the supplementary materials.



\paragraph{OptR Algorithm.}
\label{app:optr_algorithm}
Algorithm~\ref{alg:optr} summarizes the one-time offline calibration
of OptR. For each layer and KV head, OptR estimates the key mean and
optimizes only the key and value rotation corrections while keeping
all model weights frozen. The matrices $A_K$ and $A_V$ parameterize
orthogonal corrections through $\exp(A-A^\top)$. We use
$\widetilde{K}$ and $\widetilde{V}$ to denote the effective keys and
values consumed by attention after rotated INT2 quantization. Sink and
recent tokens skip INT2 quantization; their keys retain the same
centering, while their values remain unchanged in BF16.

\begin{algorithm}
\caption{OptR offline calibration}
\label{alg:optr}
\begin{algorithmic}[1]
\Require Calibration traces $\mathcal{D}$, base rotations
$R_K^0,R_V^0$, key-loss weight $\lambda_K$, optimization steps $T$
\Ensure
$\{\widehat{R}_{K,\ell,h},\widehat{R}_{V,\ell,h},
\mu_{\ell,h}\}_{\ell,h}$

\ForAll{layers $\ell$ and KV heads $h$}
    \State Estimate the key mean $\mu_{\ell,h}$ from $\mathcal{D}$
    \State Initialize $A_K,A_V\gets0$

    \For{$i=1,\ldots,T$}
        \State $R_K\gets R_K^0\exp(A_K-A_K^\top)$
        \State Construct effective keys $\widetilde{K}(R_K)$ through
        key reparameterization and rotated INT2 quantization
        \State $p_t^K\gets
        \operatorname{softmax}
        \left(
        q_{t,j}\widetilde{K}_{1:t}(R_K)^\top/\sqrt{d}
        \right)$
        \State $e_K(t,j)\gets
        \left[
        (p_t^K-p_t)^\top V_{1:t}
        \right]W_{O,j}^\top$
        \State $\mathcal{L}_K\gets
        \mathbb{E}_{(t,j)\in\mathcal{D}}
        \left[
        D_{\mathrm{KL}}(p_t\Vert p_t^K)
        +\lambda_K
        \frac{\|e_K(t,j)\|_2^2}{d_{\mathrm{model}}}
        \right]$
        \State Update $A_K$ using Adam and the INT2 STE
    \EndFor

    \State $\widehat{R}_K\gets
    R_K^0\exp(A_K-A_K^\top)$
    \State Compute $\widehat{p}_t^K$ using
    $\widetilde{K}(\widehat{R}_K)$

    \For{$i=1,\ldots,T$}
        \State $R_V\gets R_V^0\exp(A_V-A_V^\top)$
        \State Construct effective values $\widetilde{V}(R_V)$
        through rotated INT2 quantization
        \State $e_V(t,j)\gets
        \left[
        (\widehat{p}_t^K)^\top
        \left(
        \widetilde{V}_{1:t}(R_V)-V_{1:t}
        \right)
        \right]W_{O,j}^\top$
        \State $\mathcal{L}_V\gets
        \mathbb{E}_{(t,j)\in\mathcal{D}}
        \left[
        \frac{\|e_V(t,j)\|_2^2}{d_{\mathrm{model}}}
        \right]$
        \State Update $A_V$ using Adam and the INT2 STE
    \EndFor

    \State $\widehat{R}_V\gets
    R_V^0\exp(A_V-A_V^\top)$
    \State Store
    $\widehat{R}_{K,\ell,h}\gets\widehat{R}_K$ and
    $\widehat{R}_{V,\ell,h}\gets\widehat{R}_V$
\EndFor

\State \Return
$\{\widehat{R}_{K,\ell,h},
\widehat{R}_{V,\ell,h},
\mu_{\ell,h}\}_{\ell,h}$
\end{algorithmic}
\end{algorithm}

Here, $p_t$ denotes the BF16 attention distribution, $p_t^K$ is
computed using the current effective INT2 keys, and
$\widehat{p}_t^K$ is computed using the selected key rotation
$\widehat{R}_K$. The values $V_{1:t}$ remain in BF16 during key
calibration. The projection $W_{O,j}$ maps the output of query head
$j$ into the model hidden space. Layer and KV-head indices are omitted
inside the algorithm when clear.

\paragraph{Details of Figure~\ref{fig:key_error_motivation}.}We use the same 4,096-token AIME25 trace from Qwen3-8B for all five settings. The example is taken from layer 4, KV head 5, and query head 20. The top row shows the absolute key values immediately before INT2 quantization. The bottom row shows each cached key token's contribution to the post-$W_O$ error, aggregated over the final 64 query positions while keeping values in BF16 to isolate key-induced error. All panels use the same sample and INT2 configuration: group size 128, key clipping ratio 0.96, 64 BF16 sink tokens, and 256 BF16 recent tokens.

\paragraph{Calibration details.}
We collect Q/K/V activation statistics from all 198 GPQA-Diamond prompts with one generated token, without using answer labels. During collection, each dynamically formed prefill batch is stored as a chunk containing aligned Q, K, V, and sequence-length tensors, yielding 14 chunks for each Qwen model and 15 for Phi-14B. From these dumps, we use chunks 3 and 4 for optimization and chunks 10 and 12 for held-out selection. After removing sequences shorter than 328 tokens, the calibration/held-out sets contain 6/5 prompts for Qwen3-4B, 7/6 for Qwen3-8B, and 9/3 for Phi-14B, corresponding to 2,554/2,526, 2,711/3,092, and 4,151/1,619 tokens, respectively. We compute the objective over the final 64 query positions and select the best rotation for each KV head using the held-out objective every 20 steps. Key-centering statistics are estimated from 28 full model-generated traces. The calibration and held-out chunks do not overlap, although their source prompts come from the GPQA-Diamond evaluation pool.

\begin{table}[h]
\centering
\caption{Default experimental settings for OptR.}
\small
\begin{tabular}{lll}
\hline
\textbf{Setting} & \textbf{Configuration} \\
\hline
KV-cache precision
& Group-wise affine INT2 \\

Quantization group size
& 128 \\

Key clipping ratio
& 0.96 \\

Value clipping ratio
& 0.92 \\

BF16 sink window
& 64 tokens \\

BF16 recent window
& 256 tokens \\

Calibration data
& GPQA decoding traces \\

Optimizer
& Adam~\citep{kingma2014adam} \\

Optimization steps
& 80 \\

Learning rate
& 0.02 \\
\hline
\end{tabular}
\label{tab:experimental_settings}
\end{table}

\paragraph{Experimental settings.}
We implement OptR on the official OSCAR codebase and adopt its
best-performing INT2 configuration~\citep{OSCAR}. The default settings
used throughout the main experiments and appendix are summarized in
Table~\ref{tab:experimental_settings}. Unless explicitly stated, each
ablation changes only the setting under investigation while keeping all
others fixed.

We measure efficiency on a single NVIDIA A100-SXM4 40GB GPU using
Qwen3-4B-Thinking-2507 with tensor parallelism of one and an
eight-token page size. Standard CUDA graphs are enabled for each
evaluated batch size, while piecewise CUDA graphs are disabled. Each
process performs an initial untimed warm-up. For every end-to-end and
prefill setting, we additionally discard one complete warm-up run
before recording five timed runs. Decode latency is measured over
1,024 generated tokens and excludes prefill time. End-to-end output
throughput is computed as the number of generated tokens divided by
the combined prefill and decoding time. We report the mean over five
runs in the main figure and omit error bars for readability. The
maximum standard deviations across all evaluated settings are 0.104
ms per token for decode latency, 0.412 tokens/s for end-to-end
throughput, and 0.0115 s for prefill time. We use a custom SGLang
implementation, PyTorch 2.9.1 with CUDA
12.8, and Triton 3.5.1.

\paragraph{Key-mean estimation and application.}
The key mean is estimated from post-RoPE keys in the BF16 calibration
traces:
\begin{equation}
    \mu_{\ell,h}
    =
    \frac{1}{N}
    \sum_{n,s}
    k_{n,s,h}^{\ell,\mathrm{RoPE}}
\end{equation}
At inference, the same mean is subtracted from the post-RoPE keys
before rotation and quantization. The cache-write kernel implements
this operation as
\begin{equation}
    \left(k_s^{\mathrm{RoPE}}-\mu_{\ell,h}\right)R_{K,\ell,h}
    =
    k_s^{\mathrm{RoPE}}R_{K,\ell,h}
    -
    \mu_{\ell,h}R_{K,\ell,h}
\end{equation}
We precompute $\mu_{\ell,h}R_{K,\ell,h}$ once and subtract it in the
rotated space. BF16 sink and recent keys retain the same centering.

\paragraph{Rotated-query implementation.}
For clarity, the formulation in the main paper maps an effective
quantized key back with $R_K^\top$. Let
\begin{equation}
    k_{R,s}
    =
    Q_2\!\left(
    (k_s-\mu)R_K
    \right)
\end{equation}
The corresponding effective key is
$\widetilde{k}_s=k_{R,s}R_K^\top$. Its attention logit satisfies
\begin{equation}
    q_t\widetilde{k}_s^\top
    =
    q_tR_K k_{R,s}^\top
\end{equation}
Therefore, the implementation stores $k_{R,s}$ directly in the INT2
cache and applies $R_K$ to the post-RoPE query instead of explicitly
applying $R_K^\top$ to every dequantized key. Under GQA, each query
head uses the rotation of its corresponding KV head. The two
implementations produce identical attention logits.

\paragraph{Equivalent Inference Implementation.}
Figure~\ref{fig:OptR_overview} shows the baseline and output-aware
correction rotations as separate operations. We denote their combined
key and value rotations by
\begin{equation}
    R_K = R_kR_{kc}
    \qquad
    R_V = R_vR_{vc}
\end{equation}

For the key path, the same $R_K$ is applied to the post-RoPE query
and centered key. Since $R_K$ is orthogonal, the full-precision
attention logit satisfies
\begin{equation}
    (q_tR_K)
    \left((k_s-\mu)R_K\right)^\top
    =
    q_t(k_s-\mu)^\top
\end{equation}
Thus, the key rotations cancel in the full-precision inner product.
The remaining term $-q_t\mu^\top$ is constant across all cached tokens
$s$ and therefore leaves the softmax distribution unchanged.

This equivalence holds before quantization. Under INT2, the stored key
is
\begin{equation}
    k_{R,s}
    =
    Q_2\!\left((k_s-\mu)R_K\right)
\end{equation}
and the attention logit becomes $(q_tR_K)k_{R,s}^\top$
Therefore, the rotation changes the coordinate system in which the
INT2 error is introduced, although it does not change the
full-precision attention computation. This implementation is
equivalent to the inverse-rotation formulation in the main paper:
\begin{equation}
    q_t
    \left(
    k_{R,s}R_K^\top
    \right)^\top
    =
    (q_tR_K)k_{R,s}^\top
\end{equation}
We therefore store $k_{R,s}$ directly in the INT2 cache and apply
$R_K$ to each query head using the rotation of its corresponding KV
head.

For the value path, Figure~\ref{fig:OptR_overview} explicitly applies
$R_v$ and $R_{vc}$ before quantization and their inverses after
attention. With $R_V=R_vR_{vc}$, the rotated value and attention
output are
\begin{equation}
    v_s' = v_sR_V
    \qquad
    o_j' = o_jR_V
\end{equation}
The explicit inverse path recovers the original output:
\begin{equation}
    o_j'
    R_{vc}^{-1}R_v^{-1}W_{O,j}^\top
    =
    o_jW_{O,j}^\top
\end{equation}

In the implementation, these value-side transforms are folded into
the projection weights. For KV head $h$ and query head $j\in G_h$, we
use
\begin{equation}
    \overline{W}_{V,h}
    =
    R_{V,h}^\top W_{V,h}
    \qquad
    \overline{W}_{O,j}
    =
    W_{O,j}R_{V,h}
\end{equation}
These weights satisfy
\begin{align}
    x\overline{W}_{V,h}^\top
    &=
    (xW_{V,h}^\top)R_{V,h},\\
    (o_jR_{V,h})\overline{W}_{O,j}^\top
    &=
    o_jW_{O,j}^\top
\end{align}
Hence, the value rotations and their inverses shown in
Figure~\ref{fig:OptR_overview} are mathematical operations rather
than separate inference kernels.

\paragraph{Models and Benchmarks.}
We evaluate OptR across three reasoning-oriented language models that differ in
model scale and architecture. This setting allows us to examine whether the
proposed method remains effective across both compact and larger models, as well
as across distinct model families.

\begin{itemize}
    \item \textbf{Qwen3-4B-Thinking-2507 and Qwen3-8B}~\citep{qwen3}: two Qwen3 models at different scales, used to assess whether the effectiveness of OptR is preserved as model scale increases.
    \item \textbf{Phi4-14B-reasoning-plus}~\citep{phi4reasoning}: a reasoning model
    from a different model family, included to evaluate cross-architecture
    generalization.
\end{itemize}

Together, these models cover parameter scales from 4B to 14B and include both
within-family scaling and cross-family evaluation.

We evaluate reasoning, coding, and long-context retrieval capabilities to measure
the effect of INT2 KV-cache quantization across diverse inference workloads.
\begin{itemize}
    \item \textbf{AIME24 and AIME25}~\citep{aime25}: challenging mathematical
    reasoning benchmarks that require multi-step problem solving.
    \item \textbf{GPQA-Diamond}~\citep{gpqa}: a graduate-level scientific reasoning
    benchmark covering questions that require specialized knowledge and careful
    reasoning.
    \item \textbf{MBPP+}~\citep{evalplus}: a code-generation benchmark with extended
    test cases for more rigorous functional-correctness evaluation.
    \item \textbf{LiveCodeBench v6}~\citep{livecodebench}: a contamination-resistant
    coding benchmark constructed from recent programming problems.
    \item \textbf{RULER-NIAH}~\citep{ruler}: a long-context retrieval benchmark used
    to evaluate whether quantized KV caches preserve information over extended
    context lengths.
\end{itemize}

The reasoning and coding benchmarks evaluate the quality of generated outputs,
whereas RULER-NIAH isolates long-context retrieval performance under increasing
context lengths.

\noindent\textbf{Generation Settings.}
We use the same sampling configuration across all models and benchmarks to ensure
a consistent comparison. Specifically, we set the temperature to $0.6$, top-$p$
to $0.95$, and top-$k$ to $20$. For the reasoning and coding benchmarks, we use
maximum generation lengths of $32$K tokens for the Qwen3 models and $16$K tokens
for Phi4-14B-reasoning-plus.

\section{Output-Error Measurement and Analysis}
\label{app:analysis}

\begin{figure*}[t]
\centering
{\small\bfseries Qwen3-4B-Thinking - GPQA-Diamond\par}

\includegraphics[
    width=1.0\textwidth,
    height=0.175\textheight,
    keepaspectratio
]{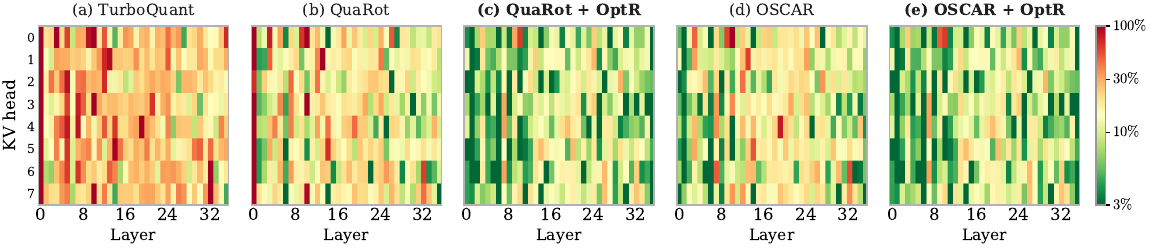}

{\small\bfseries Qwen3-8B - GPQA-Diamond\par}

\includegraphics[
    width=1.0\textwidth,
    height=0.175\textheight,
    keepaspectratio
]{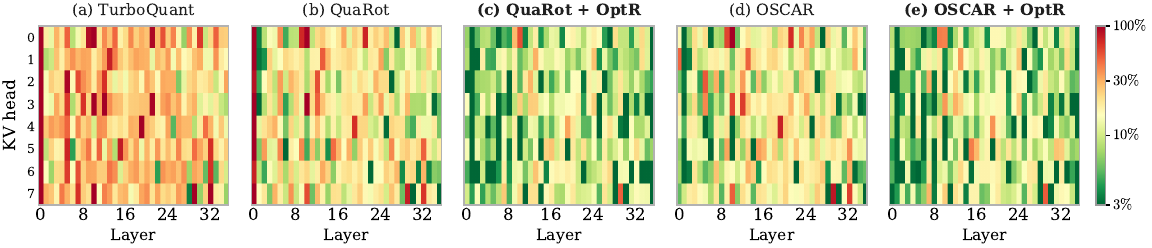}
{\small\bfseries Qwen3-4B-Thinking - AIME-25\par}

\includegraphics[
    width=1.0\textwidth,
    height=0.175\textheight,
    keepaspectratio
]{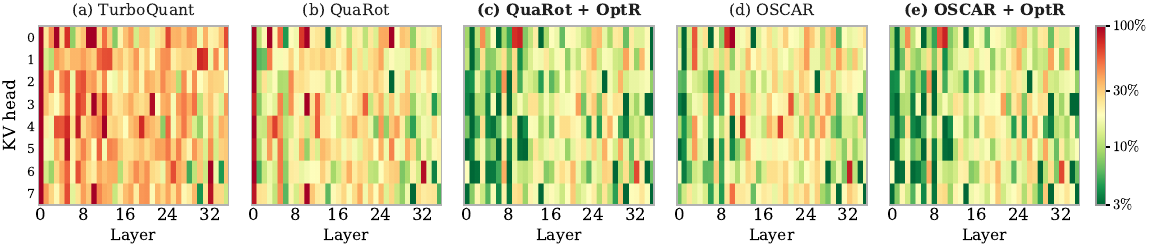}
{\small\bfseries Qwen3-8B - AIME-25\par}

\includegraphics[
    width=1.0\textwidth,
    height=0.175\textheight,
    keepaspectratio
]{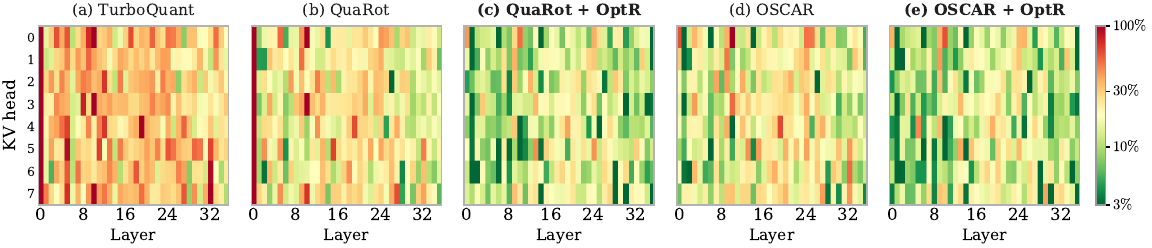}
\caption{
Output error by layer and KV head relative to naive INT2 across models and
evaluation datasets. Each cell reports the post-$W_O$ attention-output RMS
error as a percentage of plain per-group INT2 without rotation. Lower values
indicate smaller output error, with naive INT2 corresponding to $100\%$.
Within each heatmap, panels show (a) TurboQuant, (b) QuaRot,
(c) QuaRot + OptR, (d) OSCAR, and (e) OSCAR + OptR.
}
\label{fig:appendix_heatmaps}
\end{figure*}

\begin{figure*}[t]
\centering
\includegraphics[width=1.0\textwidth]{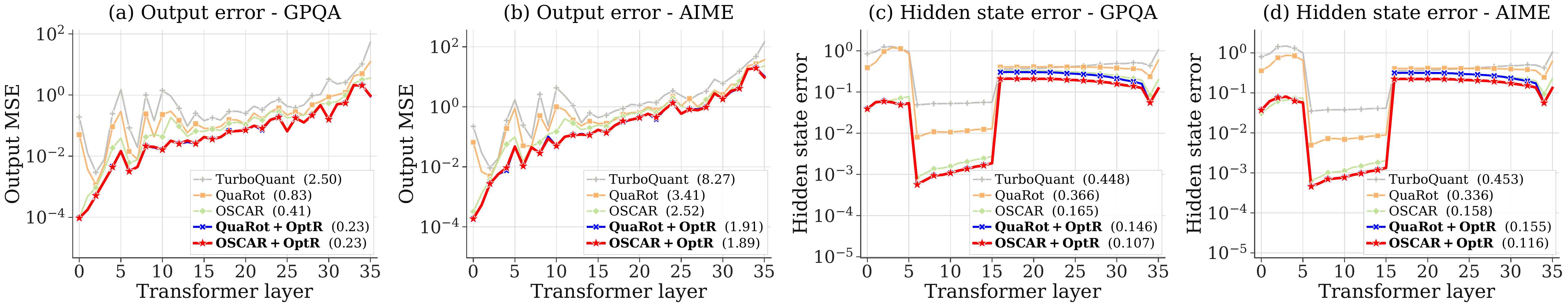}
\caption{Per-layer error analysis under INT2 KV-cache quantization on Qwen3-4B-Thinking. Panels (a,b) report the post-$W_O$ output-space MSE on GPQA-Diamond and AIME25, capturing the immediate error introduced by each attention block. Panels (c,d) report the propagated residual-stream error, computed as the normalized squared difference between full-precision and INT2-KV block-output hidden states evaluated on the same full-precision-generated sequences. Values in the legend denote averages across layers. All errors are shown on a logarithmic scale.}
\label{fig:appendix_figure1}
\end{figure*}

\begin{figure*}[t]
\centering
\includegraphics[width=1.0\textwidth]{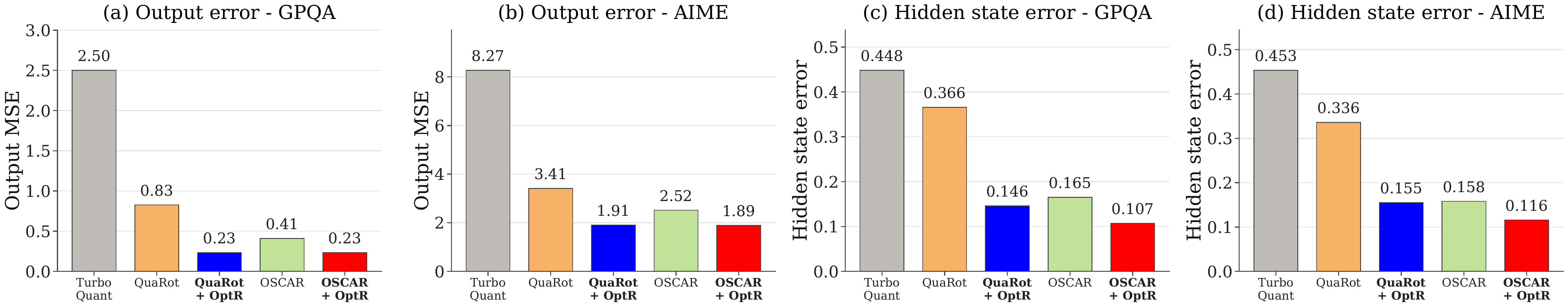}
\caption{Layer-averaged error under INT2 KV-cache quantization on Qwen3-4B-Thinking. Panels (a,b) report the mean post-$W_O$ output-space MSE across Transformer layers on GPQA-Diamond and AIME25, respectively. Panels (c,d) report the corresponding mean propagated residual-stream error. The values summarize the per-layer results shown in Figure~\ref{fig:appendix_figure1}. Lower values indicate smaller quantization-induced error.}
\label{fig:appendix_figure2}
\end{figure*}

\begin{figure*}[t]
\centering
\includegraphics[width=1.0\textwidth]{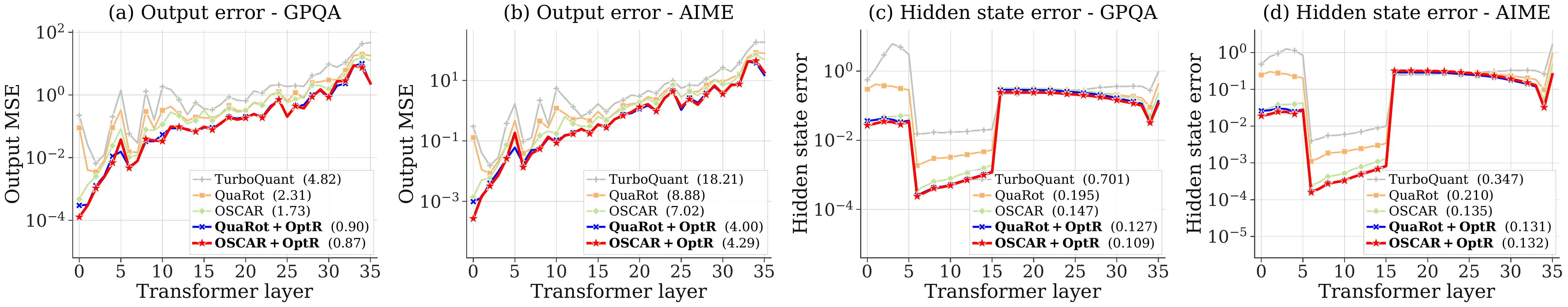}
\caption{Per-layer error analysis under INT2 KV-cache quantization on Qwen3-8B. Panels (a,b) report the post-$W_O$ output-space MSE on GPQA-Diamond and AIME25, capturing the immediate error introduced by each attention block. Panels (c,d) report the propagated residual-stream error, computed as the normalized squared difference between full-precision and INT2-KV block-output hidden states evaluated on the same full-precision-generated sequences. Values in the legend denote averages across layers. All errors are shown on a logarithmic scale.}
\label{fig:appendix_figure3}
\end{figure*}

\begin{figure*}[t]
\centering
\includegraphics[width=1.0\textwidth]{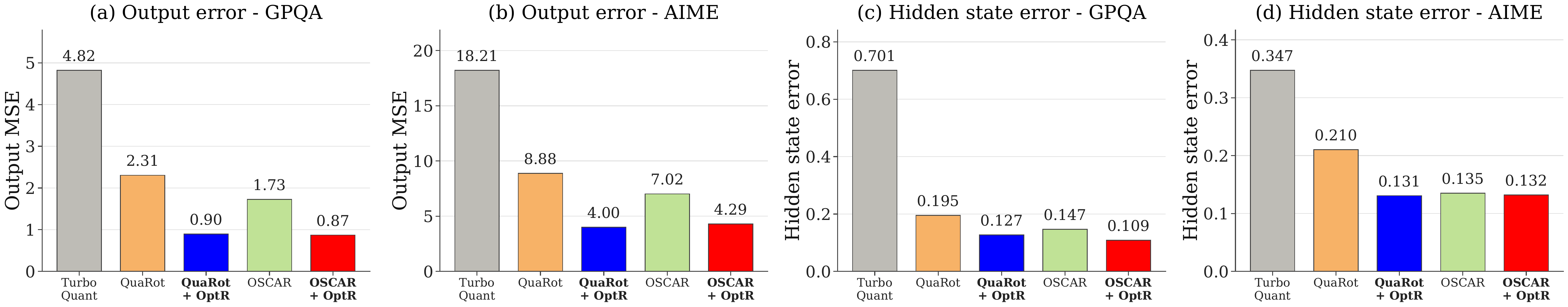}
\caption{Layer-averaged error under INT2 KV-cache quantization on Qwen3-8B. Panels (a,b) report the mean post-$W_O$ output-space MSE across Transformer layers on GPQA-Diamond and AIME25, respectively. Panels (c,d) report the corresponding mean propagated residual-stream error. The values summarize the per-layer results shown in Figure~\ref{fig:appendix_figure3}. Lower values indicate smaller quantization-induced error.}
\label{fig:appendix_figure4}
\end{figure*}

This section describes the measurement settings for the per-layer line plots,
layer-averaged bar plots, and per-head heatmaps. The main paper reports the
Qwen3-4B-Thinking-2507 per-layer analysis and GPQA-Diamond heatmap in
Figures~\ref{fig:per-layer-error} and~\ref{fig:head_heatmap}, respectively.
Additional per-layer and layer-averaged results for
Qwen3-4B-Thinking-2507 and Qwen3-8B are provided in
Figures~\ref{fig:appendix_figure1}--\ref{fig:appendix_figure4}, while the
complete per-head heatmaps across both models and datasets are shown in
Figure~\ref{fig:appendix_heatmaps}. Unless otherwise specified, all analyses
use the same INT2 quantization settings as the main experiments.

\paragraph{Output-space error.}
For each layer $\ell$, KV head $h$, and query head $j \in \mathcal{G}_h$, we
use the last $64$ query positions of each sequence. Let
$p_t^{\ell,j,h}$ and $\widetilde p_t^{\ell,j,h}$ denote the causal attention
distributions obtained with full-precision and INT2 keys, respectively. We
write $V_{1:t,h}^{\ell}$ and $\widetilde V_{1:t,h}^{\ell}$ for the
full-precision and effective INT2 value matrices, where the latter includes
full-precision sink- and recent-token restoration. Following the notation in
the main paper, the post-$W_O$ output discrepancy is
\[
  \Delta y_{t,j}^{\ell}
  =
  \left[
  \bigl(\widetilde p_t^{\ell,j,h}\bigr)^\top
  \widetilde V_{1:t,h}^{\ell}
  -
  \bigl(p_t^{\ell,j,h}\bigr)^\top
  V_{1:t,h}^{\ell}
  \right]
  \bigl(W_{O,j}^{\ell}\bigr)^\top
\]
This quantity is the total output error
$\Delta y_{t,j}^{\ell}=\widetilde y_{t,j}^{\ell}-y_{t,j}^{\ell}$ and jointly
captures key-induced changes in the attention distribution and value-induced
changes in the attention-weighted output. For each layer, we average
$\|\Delta y_{t,j}^{\ell}\|_2^2$ over query positions, query heads, KV heads,
and sequences. We apply the same group-wise INT2 quantization and
dequantization path used during inference to the KV cache of the measured
layer while using full-precision input activations. The resulting values
measure the immediate error introduced by each attention block. The
corresponding per-layer results are reported in
Figures~\ref{fig:per-layer-error}, \ref{fig:appendix_figure1},
and~\ref{fig:appendix_figure3}.

\paragraph{Propagated residual-stream error.}
To measure the accumulation of quantization error across depth, we run the
complete model with INT2 KV caches in all Transformer layers. For each of $12$
prompts per dataset, the full-precision model first generates a reasoning trace
greedily, with up to $1536$ new tokens and a maximum sequence length of $3072$.
The full-precision and INT2 models are then evaluated on the same generated
token sequence. This teacher-forced comparison isolates representational drift
from differences caused by autoregressive sampling.

All model weights remain in BF16. Long-history keys and values are
quantized and stored in INT2 at cache-write time and dequantized during
attention, while the configured sink- and recent-token windows remain in full
precision. Let $h_{\ell,t}$ and $\widetilde h_{\ell,t}$ denote the
full-precision and INT2-KV residual streams after block $\ell$ at token position
$t$. We report the relative squared error
\[
  \frac{\sum_t
  \|\widetilde h_{\ell,t}-h_{\ell,t}\|_2^2}
  {\sum_t\|h_{\ell,t}\|_2^2}
\]
pooled over all tokens and prompts. This normalization accounts for changes in
the residual-stream magnitude across depth and enables comparison across
layers. The propagated errors across model depth are shown in panels (c,d) of
Figures~\ref{fig:per-layer-error}, \ref{fig:appendix_figure1},
and~\ref{fig:appendix_figure3}.

\paragraph{Layer-averaged summaries.}
The bar plots summarize the corresponding per-layer line plots by reporting
the arithmetic mean of each error metric across Transformer layers. Panels
(a,b) report the mean post-$W_O$ output-space MSE on GPQA-Diamond and
AIME25, while panels (c,d) report the corresponding mean propagated
residual-stream error. The Qwen3-4B-Thinking-2507 and Qwen3-8B summaries are
reported in Figures~\ref{fig:appendix_figure2}
and~\ref{fig:appendix_figure4}, respectively. These plots provide an aggregate
comparison between methods, whereas
Figures~\ref{fig:appendix_figure1} and~\ref{fig:appendix_figure3} show how the
same errors vary across model depth.

\paragraph{Per-head error relative to naive INT2.}
The heatmaps use the same post-$W_O$ output error
$\Delta y_{t,j}^{\ell}$, but retain a separate value for each
$(\text{layer},\text{KV head})$ pair rather than averaging over KV heads. Let
$\mathrm{err}_{\ell,h}^{\mathrm{method}}$ denote the mean squared output error
for a given cell and let $\mathrm{err}_{\ell,h}^{\mathrm{INT2}}$ denote the
corresponding error under plain per-group INT2 without rotation or clipping.
Each cell reports
$100\sqrt{\mathrm{err}_{\ell,h}^{\mathrm{method}}/
\mathrm{err}_{\ell,h}^{\mathrm{INT2}}}$, representing the RMS output error as
a percentage of naive INT2. Thus, naive INT2 corresponds to $100\%$, and lower
values indicate smaller output error. The heatmaps use a logarithmic scale
ranging from $3\%$ to $100\%$. The GPQA-Diamond and AIME25 heatmaps  are shown in
Figures~\ref{fig:appendix_heatmaps}.

\paragraph{Results.}
Figures~\ref{fig:appendix_figure1}
and~\ref{fig:appendix_figure2} show that OptR reduces both immediate
post-$W_O$ output error and propagated residual-stream error on
Qwen3-4B-Thinking-2507. On GPQA-Diamond, OptR reduces the average
post-$W_O$ output MSE from $0.83$ to $0.23$ for QuaRot and from $0.41$ to
$0.23$ for OSCAR. On AIME25, the corresponding errors decrease from
$3.41$ to $1.91$ and from $2.52$ to $1.89$, respectively. The propagated
residual-stream error decreases from $0.366$ to $0.146$ for QuaRot and from
$0.165$ to $0.107$ for OSCAR on GPQA-Diamond, and from $0.336$ to $0.155$
and from $0.158$ to $0.116$ on AIME25.

The same trend holds for Qwen3-8B, as shown in
Figures~\ref{fig:appendix_figure3}
and~\ref{fig:appendix_figure4}. On GPQA-Diamond, OptR reduces the average
post-$W_O$ output MSE from $2.31$ to $0.90$ for QuaRot and from $1.73$ to
$0.87$ for OSCAR. On AIME25, the corresponding errors decrease from
$8.88$ to $4.00$ and from $7.02$ to $4.29$. The propagated residual-stream
error decreases from $0.195$ to $0.127$ for QuaRot and from $0.147$ to
$0.109$ for OSCAR on GPQA-Diamond. On AIME25, it decreases from $0.210$
to $0.131$ for QuaRot and from $0.135$ to $0.132$ for OSCAR.

Finally, Figure~\ref{fig:appendix_heatmaps} shows
that the error reductions extend across most layers and KV heads on both
GPQA-Diamond and AIME25. The improvements are therefore broadly
distributed throughout the models rather than being concentrated in a small
subset of layers or heads.

\begin{table}[H]
\centering
\caption{Effect of calibration size on AIME25 accuracy for
Qwen3-4B-Thinking-2507. Both methods use the same GPQA calibration tokens and
a BF16 window of $S{=}64/R{=}256$. Results are $\mu\pm\sigma$ over 5 seeds.}
\small
\begin{tabular}{lcc}
\hline
\textbf{Tokens} & \textbf{OSCAR} & \textbf{OSCAR\,+\,OptR} \\
\hline
$3.1$k  & $62.00\pm6.50$ & $69.33\pm3.65$ \\
$6.6$k  & $63.33\pm4.08$ & $70.67\pm2.79$ \\
$12.6$k & $62.67\pm5.96$ & $70.00\pm5.77$ \\
$19.2$k & $64.00\pm3.65$ & $70.67\pm4.35$ \\
$26.0$k & $62.67\pm6.41$ & $70.00\pm4.08$ \\
\hline
\end{tabular}

\label{tab:calib_size}
\end{table}

\subsection{Ablation on Calibration Size}
Table~\ref{tab:calib_size} evaluates the sensitivity of OSCAR and OptR to the
number of calibration tokens. OptR consistently improves OSCAR across all
calibration sizes, while its performance remains stable from a few thousand
tokens onward. Increasing the calibration size beyond $6.6$k tokens provides
no consistent additional gain, indicating that OptR requires only a modest
calibration set. We therefore use $6.6$k tokens as the default configuration.

\subsection{Ablation on $\lambda_K$}
Table~\ref{tab:lambda_k_ablation} studies the sensitivity of OptR to the
key-loss weight $\lambda_K$. The best value depends on the base rotation:
OSCAR + OptR performs best at $\lambda_K=1$, whereas QuaRot-INT2 + OptR
achieves its highest accuracy at $\lambda_K=10$. 

\begin{table}[h]
\centering
\caption{Effect of $\lambda_K$ on AIME25 accuracy for
Qwen3-4B-Thinking-2507. Results are $\mu\pm\sigma$ over five seeds, and the
shaded row denotes the default setting.}
\small
\begin{tabular}{ccc}
\hline
\textbf{$\lambda_K$}
& \textbf{QuaRot-INT2 + OptR}
& \textbf{OSCAR + OptR} \\
\hline

$0.1$
& $60.67 \pm 5.48$
& $66.67 \pm 4.08$ \\

$0.3$
& $62.00 \pm 4.35$
& $63.33 \pm 5.27$ \\

\rowcolor{gray!15}
$\mathbf{1}$
& $60.67 \pm 4.35$
& $\mathbf{70.67 \pm 2.79}$ \\

$3$
& $60.67 \pm 4.35$
& $63.33 \pm 4.71$ \\

$10$
& $\mathbf{63.33 \pm 4.08}$
& $60.67 \pm 2.79$ \\

\hline
\end{tabular}
\label{tab:lambda_k_ablation}
\end{table}

This difference indicates
that the appropriate balance between attention-distribution preservation and
post-$W_O$ output-error reduction depends on the rotation initialization.
To avoid base-specific tuning, we use $\lambda_K=1$ throughout the main
experiments, which gives the strongest performance with OSCAR, our
state-of-the-art rotation baseline.

\begin{table}[h]
\centering
\caption{Calibration time of OptR with 80 optimization steps. Measurements use a single NVIDIA A100 GPU and exclude BF16 trace collection. All model weights remain frozen throughout calibration.}
\small
\begin{tabular}{lc}
\hline
\textbf{Model} & \textbf{Calibration Time (Minute)} \\
\hline
Qwen3-4B-Thinking-2507 & 6.25 \\
Qwen3-8B               & 7.05 \\
Phi4-14B-Reasoning-Plus & 11.03 \\
\hline
\end{tabular}
\label{tab:calibration_cost}
\end{table}

\paragraph{Offline calibration cost.}
Table~\ref{tab:calibration_cost} reports the one-time offline
calibration cost of OptR. Calibration requires only several minutes
on a single GPU across the evaluated model sizes. All model weights
remain frozen, and the resulting rotations are reused across
subsequent inference requests, amortizing this cost during deployment.

\begin{table*}[h]
\centering
\caption{RULER-NIAH retrieval accuracy of Qwen3-8B at context lengths
up to 128K tokens. Results are $\mu\pm\sigma$ over three seeds, with
800 examples per seed. Bold denotes the best INT2 result at each
context length.}
\scriptsize
\begin{adjustbox}{width=\textwidth}
\begin{tabular}{llcccccc}
\hline
\textbf{Model} & \textbf{Method}
& \textbf{4k} & \textbf{8k} & \textbf{16k}
& \textbf{32k} & \textbf{64k} & \textbf{128k} \\
\hline

\multirow{5}{*}{Qwen3-8B}
& BF16
& $99.83 \pm 0.11$
& $99.93 \pm 0.02$
& $99.45 \pm 0.08$
& $98.70 \pm 0.49$
& $84.22 \pm 1.41$
& $80.89 \pm 0.58$ \\

& QuaRot-INT2
& $84.97 \pm 0.09$
& $41.73 \pm 5.75$
& $18.16 \pm 1.14$
& $13.03 \pm 1.71$
& $0.04 \pm 0.07$
& $0.38 \pm 0.13$ \\

& QuaRot-INT2 + \textbf{OptR}
& $99.40 \pm 0.29$
& $\mathbf{98.76 \pm 0.57}$
& $\mathbf{96.14 \pm 0.35}$
& $86.37 \pm 0.72$
& $\mathbf{70.02 \pm 1.79}$
& $\mathbf{41.87 \pm 0.96}$ \\

& OSCAR
& $99.59 \pm 0.15$
& $97.94 \pm 0.28$
& $94.39 \pm 0.29$
& $83.76 \pm 0.58$
& $57.54 \pm 1.82$
& $26.44 \pm 0.87$ \\

& OSCAR + \textbf{OptR}
& $\mathbf{99.60 \pm 0.11}$
& $98.16 \pm 0.63$
& $95.50 \pm 0.65$
& $\mathbf{86.42 \pm 0.55}$
& $68.65 \pm 0.80$
& $40.43 \pm 1.53$ \\

\hline
\end{tabular}
\end{adjustbox}
\label{tab:ruler_niah_128k_appendix}
\end{table*}

\paragraph{Long-Context Results} Table~\ref{tab:ruler_niah_128k_appendix} extends the Qwen3-8B
long-context evaluation to 128K tokens. At this context length,
QuaRot-INT2 and OSCAR degrade substantially, whereas applying OptR
retains considerably higher retrieval accuracy. The consistent gains
with both base rotations show that OptR remains effective beyond the
64K range reported in the main paper.

\end{document}